\documentclass[10pt,twocolumn,letterpaper]{article}

\usepackage{cvpr}              

\usepackage{graphicx}
\usepackage{amsmath}
\usepackage{amssymb}
\usepackage{booktabs}
\usepackage{url}
\usepackage{enumitem}
\usepackage{makecell}
\usepackage{diagbox}
\usepackage{bm}
\usepackage{dsfont}
\usepackage{multirow}
\usepackage{varwidth}
\usepackage{pifont}
\usepackage{makecell}
\usepackage{wrapfig}
\usepackage{graphicx} 
\usepackage{comment}
\usepackage{color}
\usepackage{colortbl,booktabs}
\usepackage[accsupp]{axessibility}

\usepackage{multirow}
\usepackage{footmisc}
\usepackage{color}
\usepackage[dvipsnames]{xcolor}
\usepackage{algorithm}
\usepackage{algpseudocode}
\usepackage{mathtools}
\usepackage{pifont}
\usepackage{lipsum}
\usepackage{adjustbox}
\usepackage{booktabs}       
\usepackage{amsfonts}       
\usepackage{nicefrac}       
\usepackage{microtype}      
\usepackage{makecell}
\usepackage{diagbox}

\usepackage{xspace}
\usepackage{booktabs}
\usepackage{rotating}
\usepackage{color}

\newcommand{\Mat}{\boldsymbol}
\newcommand{\Set}{\mathcal}

\newcommand{\real}{\mathbb{R}}

\DeclareMathOperator{\Prob}{\mathbb{P}}
\DeclareMathOperator{\E}{\mathbb{E}}
\DeclareMathOperator*{\Proj}{\mathrm{\Pi}}
\DeclareMathOperator*{\argmin}{arg\,min}
\DeclareMathOperator*{\argmax}{arg\,max}

\definecolor{Tianlong_color}{rgb}{0.858, 0.188, 0.478}
\definecolor{Zhiwen_color}{rgb}{0.188, 0.858, 0.478}

\usepackage[pagebackref,breaklinks,colorlinks]{hyperref}

\usepackage[capitalize]{cleveref}
\crefname{section}{Sec.}{Secs.}
\Crefname{section}{Section}{Sections}
\Crefname{table}{Table}{Tables}
\crefname{table}{Tab.}{Tabs.}

\def\cvprPaperID{4407} 
\def\confName{CVPR}
\def\confYear{2022}

\begin{document} 

\title{Aug-NeRF: Training Stronger Neural Radiance Fields with  Triple-Level \\ Physically-Grounded Augmentations}

\author{%
  Tianlong Chen\textsuperscript{1*}, Peihao Wang\textsuperscript{1*}, Zhiwen Fan\textsuperscript{1}, Zhangyang Wang\textsuperscript{1}\\
  \textsuperscript{1}The University of Texas at Austin \\
  \small{\texttt{\{tianlong.chen,peihaowang,zhiwenfan,atlaswang\}@utexas.edu}} \\
}

\maketitle

\begin{abstract}\vspace{-0.5em}
Neural Radiance Field (NeRF) regresses a neural parameterized scene by differentially rendering multi-view images with ground-truth supervision. However, when interpolating novel views, NeRF often yields inconsistent and visually non-smooth geometric results, which we consider as a \textbf{generalization gap} between seen and unseen views. Recent advances in convolutional neural networks have demonstrated the promise of advanced robust data augmentations, either random or learned, in enhancing both in-distribution and out-of-distribution generalization. Inspired by that, we propose Augmented NeRF (\textbf{Aug-NeRF}), which for the first time brings the power of robust data augmentations into regularizing the NeRF training. Particularly, our proposal learns to seamlessly blend worst-case perturbations into three distinct levels of the NeRF pipeline with physical grounds, including (1) the input coordinates, to simulate imprecise camera parameters at image capture; (2) intermediate features, to smoothen the intrinsic feature manifold; and (3) pre-rendering output, to account for the potential degradation factors in the multi-view image supervision. Extensive results demonstrate that Aug-NeRF effectively boosts NeRF performance in both novel view synthesis (up to $\mathbf{1.5} \mathrm{dB}$ PSNR gain) and underlying geometry reconstruction.
Furthermore, thanks to the implicit smooth prior injected by the triple-level augmentations, Aug-NeRF can even recover scenes from heavily corrupted images, a highly challenging setting untackled before. Our codes are available in \url{https://github.com/VITA-Group/Aug-NeRF}.\vspace{-0.5em}

\end{abstract}

\renewcommand{\thefootnote}{\fnsymbol{footnote}}
\footnotetext[1]{Equal Contribution.}
\renewcommand{\thefootnote}{\arabic{footnote}}

\section{Introduction}
\vspace{-0.5em}

\begin{figure}[t]
    \centering
    \vspace{-2mm}
    \includegraphics[width=\linewidth]{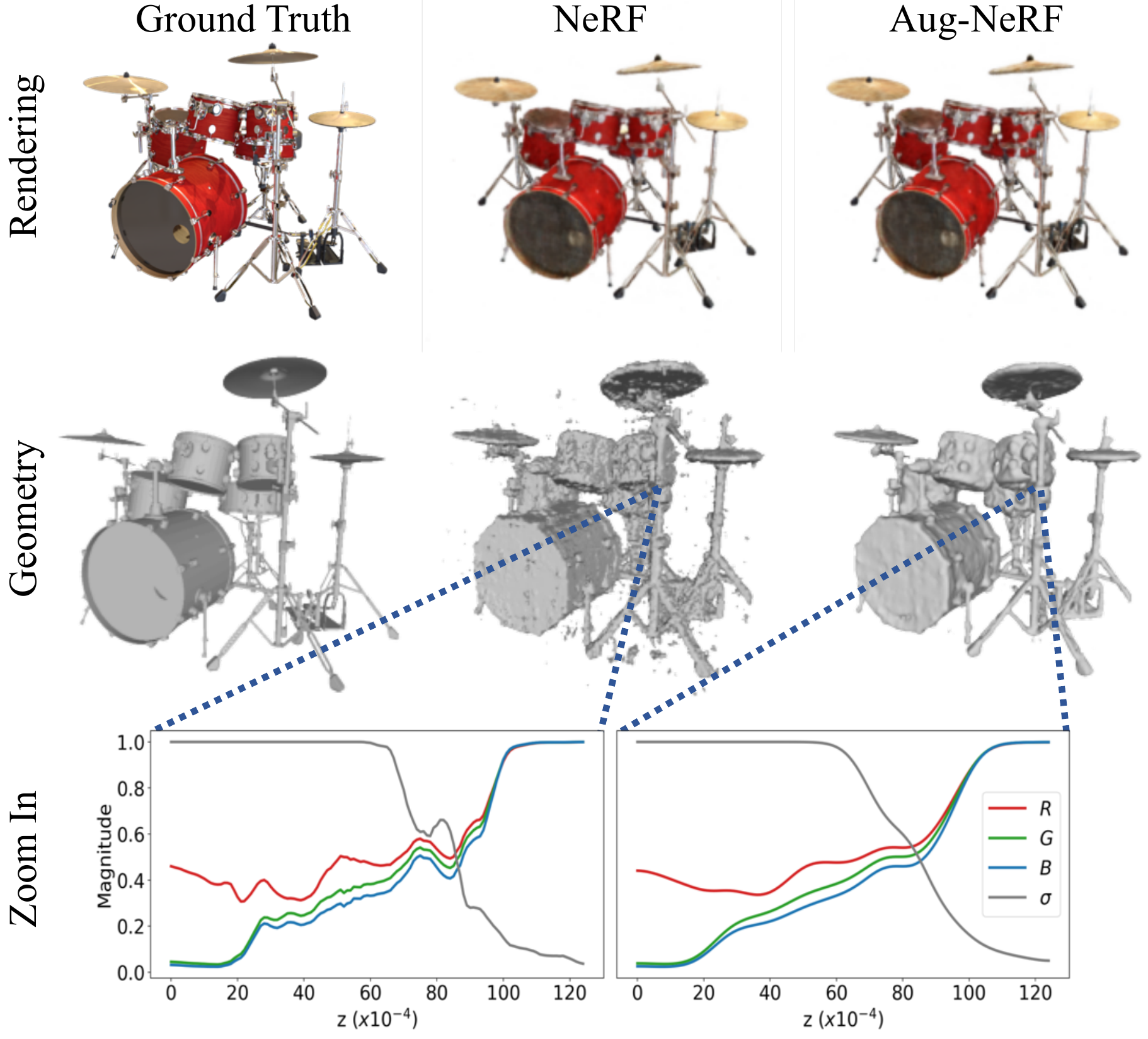}
    \vspace{-6mm}
    \caption{Comparisons between Aug-NeRF (ours) and NeRF~\cite{mildenhall2020nerf}. From upper to bottom, we present the test-set synthesized views, 3D geometry, and their zoom-in RGB$\sigma$ distributions, respectively.}
    \vspace{-3mm}
    \label{fig:teaser}
\end{figure}

Neural radiance fields (NeRF)~\cite{mildenhall2020nerf} and its variants have demonstrated impressive progresses in learning to represent 3D objects and scenes from images towards photo-realistic novel view synthesis. NeRF leverages a multi-layer perceptron (MLP) to implicitly modeling the mapping from an input 5D coordinates (i.e., 3D coordinates ($x,y,z$) and 2D viewing directions ($\theta,\phi$)) to volume density $\sigma$ and view-dependent emitted radiance color ($r,g,b$) at the corresponding position in the scene. Then, the obtained continuous 5D function (i.e., MLP) can be utilized to generate novel views with traditional volume rendering mechanisms. 

Although NeRF is capable of producing novel views, it unfortunately suffers from inconsistent and non-smooth geometries since the vanilla MLP lacks geometry-awareness. For example, as shown in Fig.~\ref{fig:teaser}, the depth maps and 3D geometries of the scene generated by NeRF show obvious discontinuity and outliers, especially around the edge of objects. Considering that the quality of reconstructed geometry plays a central role in view rendering, that might account for NeRF's limited generalization to unseen views. 

To fill in this research gap, a straightforward solution is introducing explicit geometric regularizers like Laplacian~\cite{kusupati2020normal, romanoni2019tapa} or total variation (TV)~\cite{zhou2017unsupervised} to enhance the continuity. However, these explicit regularizers are often found to constrain the representation flexibility of MLP too aggressively, resulting in inferior performance. Recent advances in robust data augmentations~\cite{xie2020adversarial} establish promising successes in image recognition in terms of both improved functional smoothness and generalization. 

Motivated by that, we design an \underline{Aug}mented NeRF (\textbf{Aug-NeRF}) training framework, which injects worst-case perturbations~\cite{madry2017towards} to implicitly regularize the NeRF pipeline with physical foundations. Specifically, Aug-NeRF considers to regularize three different levels, including ($i$) \textit{the input coordinates}, where perturbations can imitate the inaccurate camera poses during collecting images; ($ii$) \textit{the intermediate features}, in order for a smooth/flat model loss landscape~\cite{chen2021adversarial,wu2020adversarial,farnia2018generalizable} when fitting objects' 3D geometries that is believed to enhance generalization; ($iii$) \textit{the pre-rendering output}, to model potential degradation factors in the image supervision. As presented in Fig.~\ref{fig:teaser}, our Aug-NeRF achieves smoother and more consistency reconstructed geometry and improved unseen view synthesis. Additionally, we find Aug-NeRF to show surprising resilience towards severely corrupted supervision images. The main contributions of this paper can be summarized as follows:
\begin{itemize}
\vspace{-0.2em}
    \item We reveal the existence of highly non-smooth geometries in representing scenes as neural radiance fields (NeRF), which we regard as a crucial bottleneck of NeRF's generalization ability to unseen views.\vspace{-0.2em}
    \item To address such limitation of NeRF, we propose Aug-NeRF, a triple-level, physically-grounded augmented training pipeline, by leveraging worst-case perturbations to implicated regularize the input coordinate, intermediate feature, and pre-rendering output levels.\vspace{-0.2em}
    \item Extensive experiments validate the effectiveness of our proposal on diverse scene synthesis tasks, to endow NeRF with smoothness-aware geometry reconstruction, enhanced generalization to synthesizing unseen views, and stronger tolerance of noisy supervisions.\vspace{-0.2em} 
\end{itemize}

\section{Related Work}
\vspace{-0.2em}
\paragraph{Adversarial Training and Robust Augmentation.} It is well-known that deep networks are vulnerable to imperceptible worst-case perturbations~\cite{goodfellow2014explaining, kurakin2016adversarial, madry2017towards}. Numerous defense mechanisms \cite{zhang2019theoretically,schmidt2018adversarially,sun2019towards,nakkiran2019adversarial,stutz2019disentangling,raghunathan2019adversarial} have been invented to address the issue, where adversarial training (AT)  approaches~\cite{goodfellow2014explaining,kurakin2016adversarial,madry2017towards} remains as the \textit{de-facto}. Although conventional AT enhances model robustness at the price of compromising the standard accuracy~\cite{tsipras2018robustness}, recent studies reveal AT can be harnessed to enhance models' standard generalization as well ~\cite{xie2020adversarial,zhu2019freelb,wang2019improving,gan2020large,wei2019improved}. Taking~\cite{xie2020adversarial} for example, it applies adversarial perturbations to input samples as a form of data augmentation, and shows to improve image classification on the clean dataset. \cite{zhu2019freelb,wang2019improving,gan2020large} apply worst-case perturbations to the input embedding for natural language understanding, language modeling, and vision-and-language tasks, all successfully boosting their standard generalization.  \cite{zhang2019dada,hendrycks2019augmix,wang2021augmax,rebuffi2021fixing} constructed more sophisticated variations of robust augmentations, including both data-driven and heuristic components, to improve model generalization further. However, such robust augmentations on inputs or intermediate features, to our best knowledge, have not been studied in the view synthesis field. This paper explores this possibility by looking into the intrinsic physical grounds.

\vspace{-0.8em}
\paragraph{Neural 3D Representations.} 
Classic 3D reconstruction approaches utilizes discrete representations such as point clouds \cite{wu2020multi, aliev2020neural}, meshes \cite{thies2019deferred, riegler2021stable, riegler2020free}, multi-plane images \cite{14zhou2018stereo, mildenhall2019local, srinivasan2020lighthouse, srinivasan2019pushing}, depth maps~\cite{yao2018mvsnet,gu2020cascade,shrestha2021meshmvs} and voxel grids \cite{szeliski1998stereo, sitzmann2019deepvoxels, seitz1999photorealistic, penner2017soft, lombardi2019neural, kutulakos2000theory}.
Neural implicit representations leverage coordinate-based neural networks to approximate visual signals \cite{park2019deepsdf, mescheder2019occupancy, peng2020convolutional}.
Such ideas have been successfully applied to both 2D images \cite{liu2019learning, tancik2020fourier, sitzmann2020implicit} and 3D objects \cite{chibane2020implicit, saito2019pifu, sitzmann2020implicit}.
Recent advances follow differentiable rendering and end-to-end optimization to reconstruct the neural 3D scene from 2D image supervision \cite{niemeyer2020differentiable, yariv2020multiview, mildenhall2020nerf}.
Liu \emph{et al.} \cite{liu2019learning} presented the first usage of neural implicit function to infer 3D representation with differentiable rendering.
DVR \cite{niemeyer2020differentiable} and IDR \cite{yariv2020multiview} adopt surface rendering to reconstruct implicit iso-surface by supervising on both images and pixel-accurate object masks.

\begin{figure*}[t]
    \centering
    \vspace{-3mm}
    \includegraphics[width=0.90\linewidth]{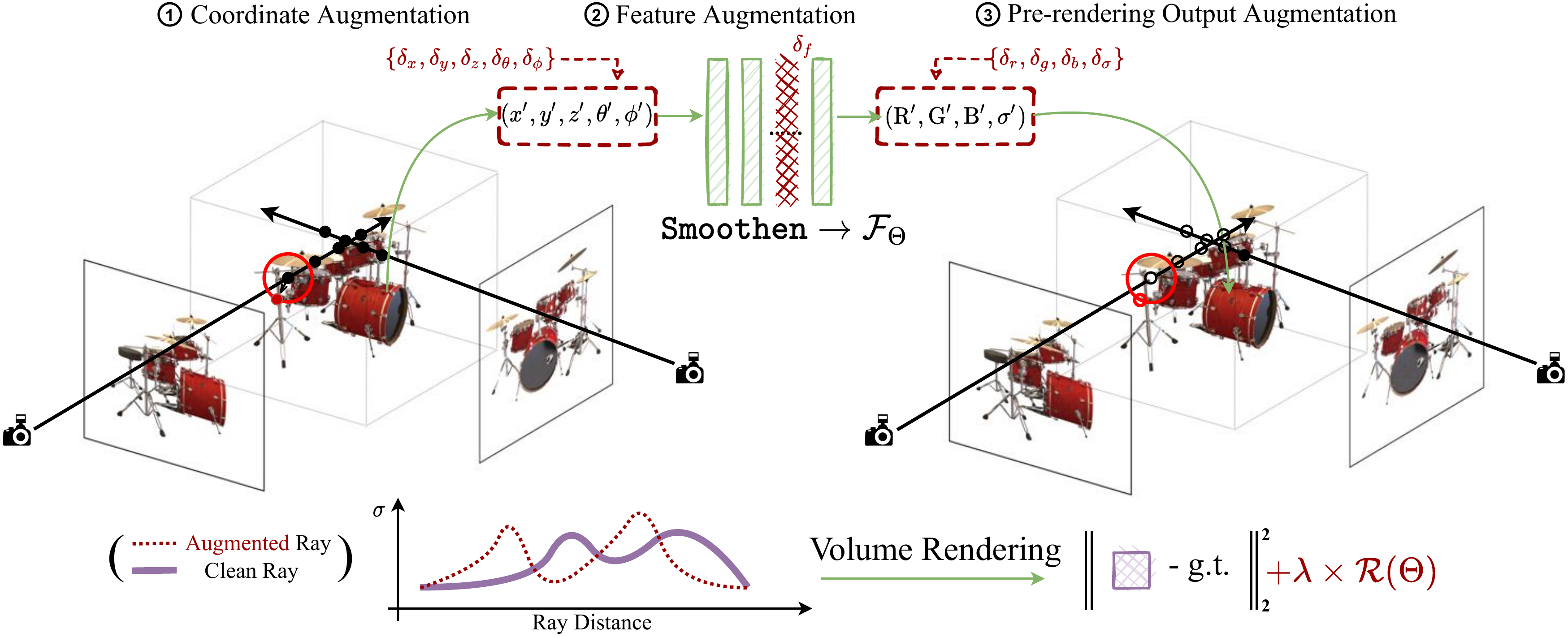}
    \vspace{-1mm}
    \caption{The overall pipeline of our proposed Aug-NeRF. The worst-case perturbations are generated at the triple levels of the NeRF pipeline: \ding{172} input coordinates, \ding{173} intermediate features, and \ding{174} pre-rendering output.}
    \vspace{-4mm}
    \label{fig:rnerf}
\end{figure*}

NeRF \cite{mildenhall2020nerf} pioneered to use differentiable volumetric rendering to optimize a neural radiance field, and achieved more photorealistic and view-consistent results. Many works continue to improve its training and rendering accuracy, efficiency, and generalization. 
NeRF++ \cite{zhang2020nerf++} separates two NeRFs to handle foreground and background, respectively. NeRF-W  \cite{martin2021nerf} tackles unstructured photos via modeling transient noises and uncertainty. MipNeRF \cite{barron2021mip} mitigates objectionable aliasing artifacts for NeRF to represent fine details. HyperNeRF \cite{park2021hypernerf} introduces topology-aware level-set methods to rectify NeRF geometry especially for dynamics. \cite{gao2020portrait, raj2021pva, rematas2021sharf, wei2021nerfingmvs} extend NeRF with lighting and rendering modeling.
\cite{oechsle2021unisurf, wang2021neus, yariv2021volume} enhance the underlying geometries reconstructed by NeRF by adopting surface representation in the place of the density volume.
\cite{wang2021ibrnet, yu2021pixelnerf, chen2021mvsnerf} leverage multi-view spatial image feature or semi-reconstructed 3D information to reduce input view number and enable generalization to new scenes.
\cite{meng2021gnerf, wang2021nerf--, yen2020inerf} free NeRF from accurate camera pose estimation.
Acceleration of NeRF training and inference have also been discussed in \cite{tancik2021learned, yu2021plenoctrees, reiser2021kilonerf}.
Despite so many exciting progresses, studying NeRF's training stability and data robustness remains an open question. 

\vspace{-0.3em}
\section{Preliminaries} \label{sec:prelim}
\vspace{-0.2em}
NeRF models the underlying 3D scene as a continuous volumetric radiance field of color and density. Formally, a typical radiance field can be written as $F: (\Mat{x}, \Mat{\theta}) \mapsto (\Mat{c}, \sigma)$, where $\Mat{x} \in \real^3$ is the spatial coordinate, $\Mat{\theta} \in [-\pi, \pi]^2$ indicates the view direction, and $\Mat{c} \in \real^{3}, \sigma \in \real_{+}$ represent the RGB color and density, respectively.
NeRF further parameterizes this 5D-valued function by a composition of Positional Embedding (PE) and the MLP $F_{\Theta} = \gamma \circ \operatorname{MLP}_{\Theta}$, where $\gamma$ is a Fourier feature mapping network~\cite{tancik2020fourier}, $\Theta$ is the network weights.
Given a radiance field, NeRF follows the classical volume rendering to render an arbitrary view~\cite{max1995optical}.

Our goal is to fit a neural radiance from calibrated RGB images captured from multiple views. 
Suppose we have a set of images with corresponding extrinsic parameters.
NeRF simulates the physical imaging process, by casting a ray $\Mat{r}=(\Mat{o}, \Mat{d}, \Mat{\theta})$ for each pixel via inverse perspective projection with respect to the camera pose, where $\Mat{o} \in \real^{3}$ denotes the optical center of camera, $\Mat{d} \in \real^{3}$ is the direction of the ray, and $\Mat{\theta} \in [-\pi, \pi]^{2}$ is the angular view direction (see Fig. \ref{fig:rnerf}). We collect all pairs of rays and pixel colors as the training set $\Set{R} = \{ (\Mat{r}_i, \widehat{\Mat{C}}_i) \}_{i=1}^{N}$, where $N$ is the total number of rays, and $\widehat{\Mat{C}}_i$ denotes the ground-truth color of the $i$-th ray.
To simulate the color of a ray, NeRF first partitions $K$ evenly-spaced bins between the near-far bound $[t_n, t_f]$ along the ray, and then uniformly samples one point within each bin: $t_k \sim \mathcal{U}[t_n + (k-1) (t_f - t_n) / K, t_f + k (t_f - t_n) / K]$. Afterwards, NeRF numerically evaluates volumetric ray integration \cite{max1995optical}  via the following equation:
\begin{align} \label{eqn:vol_render}
\Mat{C}(\Mat{r} \vert \Theta) = \sum_{k=1}^{K} T(k) (1 - \exp(-\sigma_{k} \Delta t_k)) \Mat{c}_{k} \nonumber\\
\text{where } T(k) = \exp\left( -\sum_{l=1}^{k-1} \sigma_{l} \Delta t_l \right),
\end{align}
where $\Delta t_k = t_{k+1} - t_k$, and $(\Mat{c}_{k}, \sigma_{k}) = F_{\Theta}(\Mat{o} + t_k \Mat{d}, \Mat{\theta})$. With this forward model, NeRF optimizes the expected $L_2$ distance between rendered ray colors and ground-truth pixel colors as follows:
\begin{align} \label{eqn:photo_loss}
\mathcal{L}(\Theta \vert \Set{R}) = \E_{(\Mat{r}, \widehat{\Mat{C}}) \sim \Prob(\Set{R})} \left\lVert \Mat{C}(\Mat{r} \vert \Theta) - \widehat{\Mat{C}} \right\rVert_2^2,
\end{align}
where $\Prob(\cdot)$ defines a probability measure supported in the ray space $\Set{R}$.

\section{Methodology}

\paragraph{Overview.} NeRF conducts uniform sampling along each ray and interpolates a continuous radiance field via an MLP.
However, we argue that the point sampling and the MLP interpolation can never be optimal during training dynamics due to the biased sampling strategy and non-smoothness of MLP.
To this end, we propose to train NeRF with a smoothing prior. Sec. \ref{sec:map} provides a probabilistic interpretation of this intuition.
Different from explicit smoothness modeling, e.g., total variation penalty or low rank prior, we utilize worst-case perturbations as a data-adaptive regularization. We call this training strategy Aug-NeRF. 

An overview of our Aug-NeRF is presented in Fig.~\ref{fig:rnerf}.
Following the rendering pipeline of NeRF, Aug-NeRF injects adversarial noises into the following stages: point sampling, intermediate features, and MLP outputs.
Each perturbation is searched within a small range to maximize the final loss. It could be treated as a regularization to be jointly minimized with the original training loss (see Sec. \ref{sec:advs}).

\subsection{NeRF as Maximum A Posterior} \label{sec:map}

Fitting a neural radiance field to satisfy multi-view observations can be modeled as a Maximal Likelihood (ML) problem $\Theta^* = \argmax_{\Theta} \Prob(\Set{R} | \Theta)$, which can be derived as:
\begin{align*}
\Theta^* = \argmin_{\Theta} - \mathbb{E}_{(\Mat{r}, \widehat{\Mat{C}}) \sim \Prob(\Set{R})} \log \Prob (\Mat{r}, \widehat{\Mat{C}} \big\vert \Theta),
\end{align*}
by assuming each ray is conditionally independent given network parameters.
Optimizing NeRF by MSE loss (Eqn. \ref{eqn:photo_loss}) can be obtained by regarding the conditional distribution $\Prob(\Mat{r}, \widehat{\Mat{C}} \big\vert \Theta)$ as a Gaussian distribution:
\begin{align*}
\Prob( \Mat{r}, \widehat{\Mat{C}} \big\vert \Theta ) = \frac{1}{Z} \exp \left( -\frac{1}{2\Sigma} \left\lVert \Mat{C}(\Mat{r} \vert \Theta) - \widehat{\Mat{C}} \right\rVert_2^2 \right), 
\end{align*}
where $Z$ is a normalization term, and $\Sigma$ is the variance.

However, the maximum likelihood does not introduce any prior on the reconstructed NeRF as MLP is a universal approximator.
Instead, we consider the Maximum A Posterior (MAP) form $\Prob(\Theta | \Set{R})$ to inject the prior for robust training. By Bayesian rule, we have $\Prob(\Theta | \Set{R}) \propto  \Prob(\Set{R} | \Theta) \Prob(\Theta)$, where $\Prob(\Theta)$ is some prior distribution of the network weights $\Theta$. Hence, maximizing this posterior probability is equivalent to minimizing the original loss (Eqn. \ref{eqn:photo_loss}) plus a penalty term:
\begin{align}
    \mathcal{L}(\Theta \vert \Set{R}) = \E_{(\Mat{r}, \widehat{\Mat{C}}) \sim \Prob(\Set{R})} \left\lVert \Mat{C}(\Mat{r} \vert \Theta) - \widehat{\Mat{C}}\right\rVert_2^2 + \lambda R(\Theta),
\end{align}
where $R(\Theta) = -\log \Prob(\Theta) / \lambda$. Here we expect $\Theta$ to induce a geometry-aware smooth $F_{\Theta}$.

\subsection{Regularize NeRF with Robust Augmentations} \label{sec:advs}

Imposing smoothness onto NeRF can be done in many explicit ways, such as regularizing total variation \cite{zhou2017unsupervised}, Laplacian of surface~\cite{sorkine2004laplacian, yifan2021iso, cremers2007review}, etc. However, those regularizers are often not sufficiently data-adaptive, and can constrain the representation flexibility too aggressively, as evidenced in Sec.~\ref{sec:ablation}. Also, their computation also usually operates on discretized volumetric representations, and needs extra differentiation steps to be added in NeRF.

Recent works~\cite{xie2020adversarial,zhu2019freelb,wang2019improving,gan2020large,wei2019improved,chen2021adversarial} suggest a promising alternative by integrating worst-case adversarial perturbations as data augmentations (i.e., AT). AT restricts the change of loss when its input is perturbed, leading to flattening the loss landscape \cite{miyato2018virtual,stutz2021relating}. As a result, the trained network's intrinsic feature manifold and loss landscape become smoother. Prevailing theories \cite{neyshabur2017exploring,li2018visualizing,jiang2019fantastic} link the generalization ability of deep networks to the geometry of the loss landscape; in particular, a model trained to converge to wide valleys (i.e., flat basins) in loss landscape shows better generalization ability as well as robustness to distributional shifts. 

NeRF is trained by given 2D image views (often with known camera poses) and is tested to synthesize novel views from unseen angles. Intuitively, the unsatisfactory novel view synthesis could be seen as a training-testing ``generalization gap" issue. This inspires us to incorporate robust augmentations into NeRF to induce a data-adaptive smoothness prior that enhances generalization.

Designing dedicated perturbations for NeRF is far from trivial due to its inherent physics. Unlike conventional deep models, the forward pass of NeRF consists of two white-box simulating stages (point sampling, volumetric rendering) and one black-box network mapping stage.
We propose to inject worst-case perturbations into all three levels: coordinates, intermediate features of MLP, and pre-rendering MLP output: all with clear physical meanings. Formally, our approach can be formulated as a min-max game:
\begin{align} \label{eqn:min_max}
& \min_{\Theta} \E_{(\Mat{r}, \widehat{\Mat{C}}) \sim \Prob(\Set{R})} \max_{\Mat{\delta}} \left\lVert \Mat{C^\dag}(\Mat{r} \vert \Theta, \Mat{\delta}) - \widehat{\Mat{C}} \right\rVert_2^2, \nonumber\\
& \text{where } \Mat{\delta} = (\Mat{\delta}_{p}, \Mat{\delta}_{f}, \Mat{\delta}_{r}) \in \Set{S}_p \times \Set{S}_f \times \Set{S}_r,
\end{align}
where $\Mat{\delta}_{p}$, $\Mat{\delta}_{f}$, and $\Mat{\delta}_{o}$ are the perturbations to be learned and injected to the input coordinate, intermediate MLP feature, and pre-rendering RGB-$\sigma$ output, respectively, where $\Set{S}_p \subseteq \real^{6}$, $\Set{S}_f \subseteq \real^{D}$, and $\Set{S}_{r} \subseteq \real^{4}$ are the corresponding perturbation search range, $D$ is the hidden dimension of the MLP. We elaborate on each perturbation as below.

\begin{figure*}[t]
    \centering
    \vspace{-2mm}
    \includegraphics[width=0.85\linewidth]{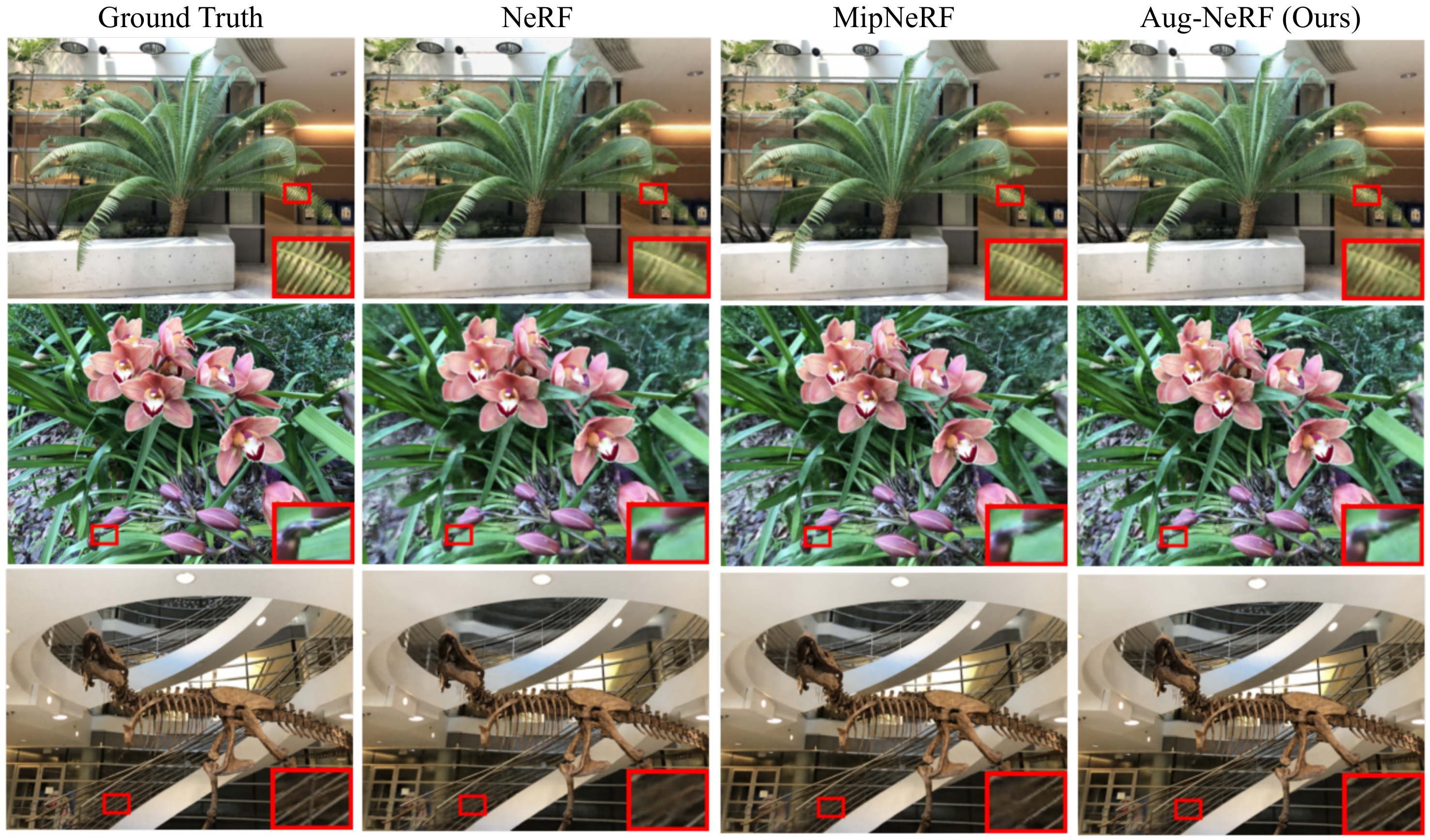}
    \vspace{-3mm}
    \caption{Comparisons on the test-set views for scenes from the realistic LLFF dataset ~\cite{mildenhall2020nerf}: local zoom-in in the \textcolor{red}{red} box.}
    \vspace{-4mm}
    \label{fig:res_llff}
\end{figure*}

\vspace{-1em}
\paragraph{Input Coordinate Perturbation.}
The original NeRF first randomly samples point along each ray and then conducts importance sampling to simulate the quadrature of the integration. This strategy also mitigates overfitting and produces smoother scene representation \cite{mildenhall2020nerf}.
Arandjelovic \textit{et al.} \cite{arandjelovic2021nerf} further proposes an attention-guided sampling scheme to refine this process.
However, our insight is that using either coarse-to-fine or learning-based sampling will cause the sampling to overfit the density distribution of the currently rendered ray, which might hold back NeRF when the density field is biased or cannot generalize.

To this end, we propose to produce a worst-case point sampling during training, to simulate a test-time ``distributional shift" for NeRF to handle.
To be specific, we search a coordinate perturbation $\Mat{\delta}_{xyx}=(\delta_x,\delta_y,\delta_z)$ following Eqn.~\ref{eqn:min_max}.
The coordinate perturbation $\Mat{\delta}_{p} = (\delta_t, \Mat{\delta}_{xyz}, \Mat{\delta}_{\theta} )^T$ consists of three parts:
1) the along-ray perturbation $\delta_t \in \real$ shifts point samples along the ray,
2) the point position perturbation $\Mat{\delta}_{xyz} \in \real^3$ is added to the direct input of the NeRF MLP,
3) in addition, we also inject the perturbation $\Mat{\delta}_{\theta} \in \real^3$ to the view direction. Formally, given the perturbation $\Mat{\delta}_{p}$, the input of MLP turns out to be:
\begin{align*}
    t^\dag_k = t_k + \delta_t, \hspace{0.5em}
    \Mat{\theta^\dag} = \Mat{\theta} + \Mat{\delta}_{\theta}, \hspace{0.5em}
    \Mat{p^\dag}_{k} = \Mat{o} + t^\dag_k \Mat{d} + \Mat{\delta}_{xyz}.
\end{align*}
The constraint set for $\delta_t$ is defined as $\delta_t \le \lvert\alpha_t (t_{k+1} - t_k)\rvert$, where $\alpha_t$ is a hyperparamter.
The coordinate perturbation $\Mat{\delta}_{xyz}$ lies in a ball $\Set{B}(0, \epsilon_p)$ to constrain points with a cylinder along the ray.
View direction perturbation $\Mat{\delta}_\theta$ is restricted within the conical frustum $[-\epsilon_p/2f, \epsilon_p/2f]^2$, where $f$ is the focal length, and $\epsilon_p$ is the pixel size.

\vspace{-1em}
\paragraph{Pre-Rendering Output Perturbation.}
NeRF next maps points on a ray to the corresponding color and density, then conducts volumetric rendering to compose these point values into the 2D pixel values. As shown by Fig. \ref{fig:teaser}, the reconstructed shape can be noisy and discontinuous. We attribute these artifacts to two reasons:
\textit{(i)} neural implicit functions represented by MLP are not necessarily smooth~\cite{tancik2020fourier, yang2021geometry}. When zooming in, we observe the function landscape to be rugged; \textit{(ii)} the MLP output goes through volumetric rendering to form the RGB output. As the volumetric rendering itself has smoothing effects owing to its point-by-point accumulation, it might ``mask" the non-smoothness and noise of the pre-rendering results hence they cannot be effectively eliminated at supervised training.

Inspired by robust training enhancing output smoothness \cite{xie2020adversarial,zhu2019freelb,wang2019improving,gan2020large,wei2019improved,chen2021adversarial}, we propose to intentionally corrupt the output of the MLP with worst-case pertubation, in order to encourage the output smoothness of the MLP, which in turn smooths the NeRF underlying geometry. Given the pre-rendering perturbation $\Mat{\delta}_{r} = (\Mat{\delta}_{c}, \delta_{\sigma})$, $\Mat{\delta}_{c} = (\delta_r,\delta_g,\delta_b)$, we perturb the rendering in Eqn. \ref{eqn:vol_render} by:
\begin{align*}
& \Mat{C^\dag}(\Mat{r} \vert \Theta) = \sum_{k=1}^{K} T(k) (1 - \exp(-(\sigma_{k} + \delta_{\sigma}) \Delta t^\dag_k)) (\Mat{c}_{k} + \Mat{\delta}_c),
\end{align*}
where $(\Mat{c}_{k}, \sigma_{k}) = F_{\Theta}(\Mat{p^\dag}_{k}, \Mat{\theta^\dag})$ are outputs by perturbed coordinates, $T(k) = \exp\left( -\sum_{l=1}^{k-1} (\sigma_{l} + \delta_{\sigma}) \Delta t^\dag_l \right)$ is the transmittance term, $\Delta t^\dag_k = t^\dag_{k+1} - t^\dag_k$ is the interval of integral, and $\Mat{\delta}_{c}, \delta_{\sigma}$ correspond to color and density perturbations, respectively.
We fix the constraint set as $[-\epsilon_c, \epsilon_c]^3 \times [-\epsilon_{\sigma}, \epsilon_{\sigma}]$.
$\Mat{\delta}_{c}, \delta_{\sigma}$ will be further clamped to make sure $\Mat{c}_{k}, \sigma_{k}$ lie between $[0, 1]$.

\vspace{-1em}
\paragraph{Intermediate Feature Perturbation.}
In addition to perturbing per-rendering color and ray density, we also inject adversarial noise into the intermediate features. As revealed by~\cite{chen2021adversarial}, augmenting intermediate features can further smooth learned functional mappings, more than just augmenting inputs or outputs. To be specific, according to \cite{mildenhall2020nerf}, the backbone MLP can be written as $(\Mat{c}(\Mat{p}), \sigma(\Mat{p})) = F_\Theta(\Mat{p}, \Mat{\theta}) = (g \circ f(\Mat{p}, \Mat{\theta}), h \circ f(\Mat{p}))$, where $f(\cdot)$ (with positional encoding) maps a coordinate to a $D$-dimension feature vector, and $g(\cdot), h(\cdot)$ project it to RGB color and density, respectively.
We crafted the worst-case perturbations as follows:
\begin{align*}
\Mat{c}_{k} = g(f(\Mat{p^\dag}_{k}) + \Mat{\delta}_f, \Mat{\theta^\dag}), \hspace{1em}
\sigma_{k} = h(f(\Mat{p^\dag}_{k}) + \Mat{\delta}_f).
\end{align*}
Intermediate feature perturbation is searched over $\Set{S}_f = [-\epsilon_f, \epsilon_f]^D$ with a hyperparameter $\epsilon_f$. We also test various injection points of the backbone MLP in Sec. \ref{sec:ablation}.

\subsection{Optimization}

To search for the worst-case perturbation in Eqn. \ref{eqn:min_max}, we introduce a theoretically guaranteed way to reach the maximum.
We only consider additive perturbation here, and all search spaces (i.e., $\Set{S}_p, \Set{S}_f, \Set{S}_r$) are defined as $\ell_p$ norm ball with a radius $\epsilon > 0$. The radius $\epsilon$ is the maximum magnitude of the perturbation, which can roughly signify the strength of the perturbation.
The perturbations can be accurately estimated by multi-step Projected Gradient Descent (PGD). Taking $\ell_{\infty}$ norm ball for example:
\begin{align} \label{eqn:pgd}
\Mat{\delta}^{(t+1)} = \Proj_{\lVert \Mat{\delta} \rVert_\infty \le \epsilon} \left[ \Mat{\delta}^{(t)} + \alpha \cdot \operatorname{sgn}(\nabla_{\Mat{\delta}} \mathcal{L}(\Theta \vert \Set{R}, \Mat{\delta}) \right]
\end{align}
where $\alpha$ is the step size of the inner maximization, $\Proj[\cdot]$ denotes a projection operator, $\operatorname{sgn}(\cdot)$ takes the sign of the input, and $\mathcal{L}(\Theta \vert \Set{R}, \Mat{\delta})$ represents the MSE loss between perturbed color $\Mat{C^\dag}$ and ground-truth color $\widehat{\Mat{C}}$ (see Eqn. \ref{eqn:min_max}).

After incorporating all augmentations, the full training objective is defined as ($\lambda = 0.1$ as tuned by grid search):
\begin{align*}
\E_{(\Mat{r}, \widehat{\Mat{C}}) \sim \Prob(\Set{R})} \underbrace{\left\lVert \Mat{C}(\Mat{r} \vert \Theta) - \widehat{\Mat{C}} \right\rVert_2^2}_{\text{photometric loss}} +
\lambda \underbrace{\max_{\Mat{\delta}} \left\lVert \Mat{C^\dag}(\Mat{r} \vert \Theta, \Mat{\delta}) - \widehat{\Mat{C}} \right\rVert_2^2}_{\text{adversarial reg.}}.
\end{align*}

\begin{figure}[t]
    \centering
    \includegraphics[width=0.95\linewidth]{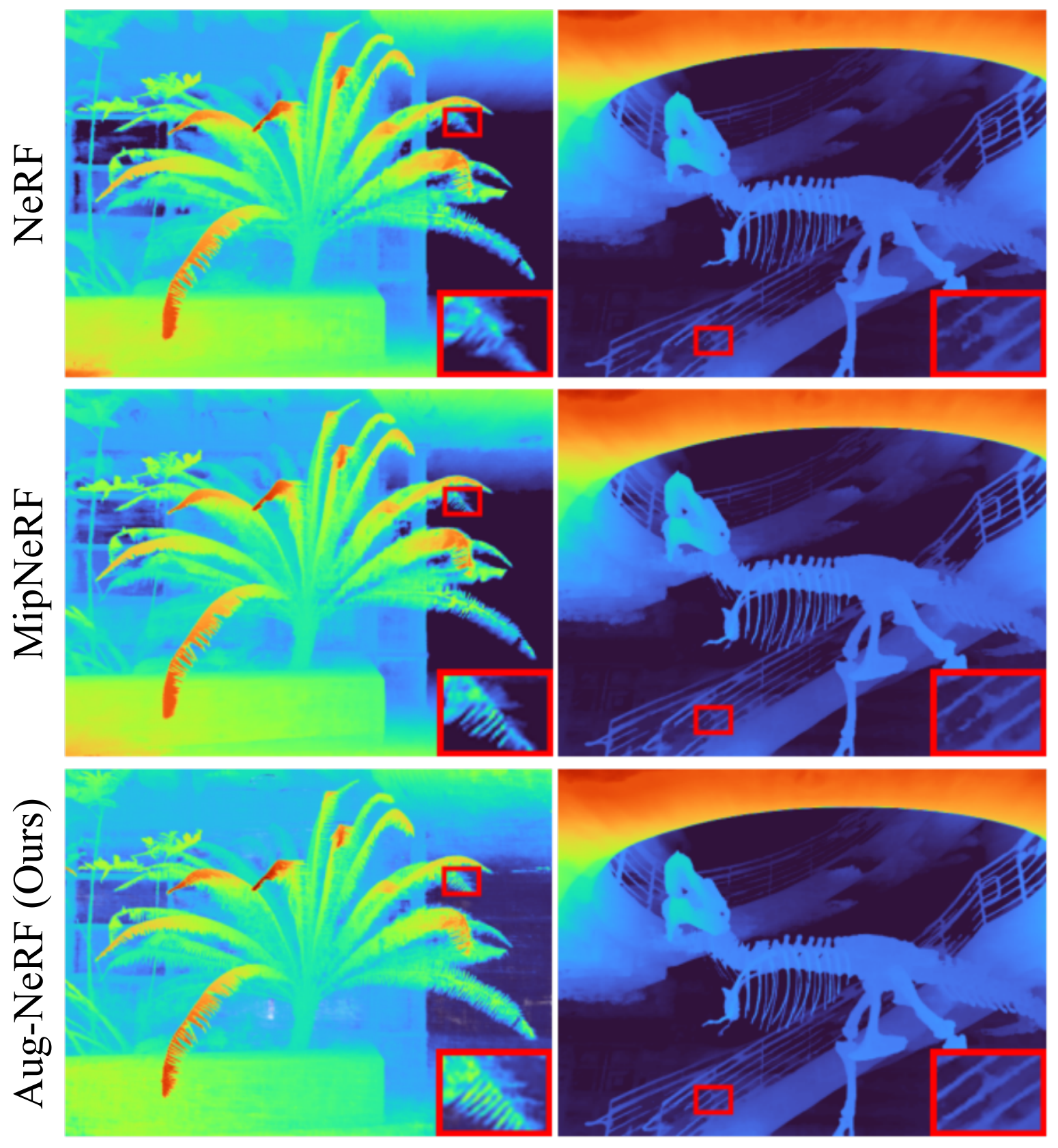}
    \vspace{-3mm}
    \caption{Comparisons of learned depth maps on scenes from LLFF dataset. The local zoom-in is placed in the \textcolor{red}{red} box.}
    \vspace{-3mm}
    \label{fig:depth}
\end{figure}

\section{Experiments}

\subsection{Implementation details.} 

\textit{Datasets.} We evaluate our proposals on public representative datasets of both LLFF~\cite{mildenhall2019local} and NeRF-Synthetic~\cite{mildenhall2020nerf}. Particularly, the face-forwarding scenes \{``fern", ``orchids", ``trex"\} from LLFF dataset and \{``drums", ``ship", and ``chair"\} instances in 360\textdegree~NeRF-Sythetic dataset are adopted in our experiments.
To accelerate training, we down-sampled LLFF dataset by 1/8 and 360\textdegree~NeRF-Synthetic dataset by 1/2.

\textit{Training.} We employ the same MLP architecture and training recipe with the original NeRF. Aug-NeRF is trained for 500K iterations to guarantee convergence.
All hyperparameters are carefully tuned by a grid search and the best configuration is applied to all experiments, as demonstrated in Sec.~\ref{sec:ablation}. NeRF models are trained on a NVIDIA RTX A$6000$ GPU with $48$ GB memory.

\textit{Evaluation.} We report three error metrics including peak signal-to-noise ratio (PSNR), the structural similarity index measure (SSIM)~\cite{1284395}, and learned perceptual image patch similarity (LPIPS)~\cite{zhang2018unreasonable}. Meanwhile, to provide a comprehensive comparison, we also follow~\cite{barron2021mip} and show an ``average" error metric by computing the geometric mean of $\mathrm{MSE=10^{\frac{-\mathrm{PSNR}}{10}}}$, $\sqrt{1-\mathrm{SSIM}}$, and $\mathrm{LPIPS}$.

\textit{Baseline and Comparison Variants.} Our Aug-NeRF is established on the vanilla NeRF~\cite{mildenhall2020nerf}. Two groups of current top-performers for view synthesis are compared, including ($i$) NeRF-based approaches: NeRF~\cite{mildenhall2020nerf} and MipNeRF~\cite{barron2021mip}; and ($ii$) classical methods: Neural Volume (NV)~\cite{lombardi2019neural}, Scene Representation Network (SRN)~\cite{sitzmann2020implicit}, and Local Light Field Fusion (LLFF)~\cite{mildenhall2019local}. For a fair comparison, all above models are trained/tested on the same views of identical scenes.

\subsection{Improved NeRF with Augmentations}

\begin{table}[t]
\caption{Quantitative comparison of our Aug-NeRF against NeRF and other top-performers for novel view synthesis. Performance is reported on the LLFF test set. $\uparrow$/$\downarrow$ means that larger/smaller numbers denote better performance.}
\label{tab:res_llff}
\centering
\vspace{-3mm}
\resizebox{0.9\linewidth}{!}{
\begin{tabular}{l|ccc|c}
\toprule
\multirow{1}{*}{Scene ``\textit{fern}"} & \multirow{1}{*}{PSNR $\uparrow$} & \multirow{1}{*}{SSIM $\uparrow$} & \multirow{1}{*}{LPIPS $\downarrow$} & \multirow{1}{*}{Average $\downarrow$}\\ 
\midrule
SRN~\cite{sitzmann2020implicit} & 21.37 & 0.611 & 0.459 & 0.128 \\
LLFF~\cite{sitzmann2020implicit} & 22.85 & 0.753 & 0.247 & 0.086 \\ \midrule
NeRF~\cite{mildenhall2020nerf} & 25.17 & 0.792 & 0.280 & 0.073 \\
MipNeRF~\cite{barron2021mip} & 26.24 & \textbf{0.839} & 0.193 & 0.057 \\ \midrule
Aug-NeRF (Ours) & \textbf{26.51} & 0.830 & \textbf{0.168} & \textbf{0.054}\\
\bottomrule
\end{tabular}}
\resizebox{0.9\linewidth}{!}{
\begin{tabular}{l|ccc|c}
\toprule
\multirow{1}{*}{Scene ``\textit{orchids}"} & \multirow{1}{*}{PSNR $\uparrow$} & \multirow{1}{*}{SSIM $\uparrow$} & \multirow{1}{*}{LPIPS $\downarrow$} & \multirow{1}{*}{Average $\downarrow$}\\  
\midrule
SRN~\cite{sitzmann2020implicit} & 17.37 & 0.611 & 0.467 & 0.175\\
LLFF~\cite{mildenhall2019local} & 18.52 & 0.588 & 0.313 & 0.141\\ \midrule
NeRF~\cite{mildenhall2020nerf} & 20.36 & 0.641 & 0.321 & 0.121\\ 
MipNeRF~\cite{barron2021mip} & 20.87 & 0.663 & 0.262 & 0.108 \\ \midrule
Aug-NeRF (Ours) & \textbf{21.60} &  \textbf{0.675} & \textbf{0.243} & \textbf{0.099}\\
\bottomrule
\end{tabular}}
\resizebox{0.9\linewidth}{!}{
\begin{tabular}{l|ccc|c}
\toprule
\multirow{1}{*}{Scene ``\textit{trex}"} & \multirow{1}{*}{PSNR $\uparrow$} & \multirow{1}{*}{SSIM $\uparrow$} & \multirow{1}{*}{LPIPS $\downarrow$} & \multirow{1}{*}{Average $\downarrow$}\\  
\midrule
SRN~\cite{sitzmann2020implicit} & 22.87 & 0.761 & 0.298 & 0.091\\
LLFF~\cite{mildenhall2019local} & 24.15 & 0.857 & 0.222 & 0.069\\ \midrule
NeRF~\cite{mildenhall2020nerf} & 26.80 & 0.880 & 0.249 & 0.056\\
MipNeRF~\cite{barron2021mip} & 27.55 & \textbf{0.894} & 0.208 & 0.049 \\ \midrule
Aug-NeRF (Ours) & \textbf{28.17} & 0.881 & \textbf{0.206} & \textbf{0.048} \\
\bottomrule
\end{tabular}}
\vspace{-3mm}
\end{table}

\paragraph{Results on LLFF and 360\textdegree~NeRF-Sythetic datasets.} In this section, we validate our proposed Aug-NeRF on LLFF and 360\textdegree~NeRF-Sythetic datasets across six representative scenes. Quantitative comparisons against vanilla NeRF and other top-performing algorithms like \{MipNeRF~\cite{barron2021mip}, NV~\cite{lombardi2019neural}, SRN~\cite{sitzmann2020implicit}, LLFF~\cite{mildenhall2019local}\} are provided in Tab.~\ref{tab:res_llff} and~\ref{tab:res_360}, together with qualitative test views presented in Fig.~\ref{fig:res_llff}. These results convey several observations:
\begin{itemize}\vspace{-0.3em}
    \item [\ding{172}] Aug-NeRF reduces average error by $14.3\%\sim 26.0\%$ and $12.5\%\sim 44.7\%$ on the LLFF and 360\textdegree~NeRF-Sythetic datasets, respectively. It consistently outperforms NeRF on all metrics by a large margin, e.g., \{$1.34$, $1.24$, $1.37$, $1.33$, $0.53$, $0.87$\} PSNR improvements at scenes \{``fern", ``orchids", ``trex", ``drums", ``ship", ``chair"\}, showing impressive ``generalization" boosts on unseen views thanks to our augmentations.\vspace{-0.2em} 
    \item [\ding{173}] Compared with recent state-of-the-art MipNeRF and other classical approaches, Aug-NeRF shows a clear advantage, especially in terms of PSNR. In some cases, MipNeRF has a slightly higher SSIM; but Aug-NeRF is able to outperform it in most cases.\vspace{-0.2em}
    \item [\ding{174}] Aug-NeRF achieves superior performance in representing fine geometry, as shown in Fig.~\ref{fig:res_llff} such as Fern's and Orchid's leaves, the skeleton ribs, and railing in T-rex. Both NeRF and MipNeRF reconstruct the low-frequency geometry and color variation, but fail to generate high-quality fine details (see zoom-in).\vspace{-0.3em}
\end{itemize}

\begin{table}[t]
\caption{Quantitative comparison of our Aug-NeRF against NeRF and other top-performer for novel view synthesis. Performance is reported on the test set of 360\textdegree~NeRF-Sythetic dataset. $\uparrow$/$\downarrow$ means that larger/smaller numbers denote better performance.
}
\label{tab:res_360}
\centering
\vspace{-3mm}
\resizebox{0.9\linewidth}{!}{
\begin{tabular}{l|ccc|c}
\toprule
\multirow{1}{*}{Scene ``\textit{drums}"} & \multirow{1}{*}{PSNR $\uparrow$} & \multirow{1}{*}{SSIM $\uparrow$} & \multirow{1}{*}{LPIPS $\downarrow$} & \multirow{1}{*}{Average $\downarrow$}\\  
\midrule
SRN~\cite{sitzmann2020implicit} & 17.18 & 0.766 & 0.267 & 0.135\\
NV~\cite{lombardi2019neural} & 22.58 & 0.873 & 0.214 & 0.075\\
LLFF~\cite{sitzmann2020implicit} & 21.13 & 0.890 & 0.126 & 0.069\\ \midrule
NeRF~\cite{mildenhall2020nerf} & 25.01 & 0.925 & 0.091 & 0.043\\
MipNeRF~\cite{barron2021mip} & 26.22 & 0.939 & 0.065 & \textbf{0.034} \\ \midrule
Aug-NeRF (Ours) & \textbf{26.34} & \textbf{0.941} & \textbf{0.060} & \textbf{0.032}\\
\bottomrule
\end{tabular}}
\resizebox{0.9\linewidth}{!}{
\begin{tabular}{l|ccc|c}
\toprule
\multirow{1}{*}{Scene ``\textit{ship}"} & \multirow{1}{*}{PSNR $\uparrow$} & \multirow{1}{*}{SSIM $\uparrow$} & \multirow{1}{*}{LPIPS $\downarrow$} & \multirow{1}{*}{Average $\downarrow$}\\
\midrule
SRN~\cite{sitzmann2020implicit} &  20.60 & 0.757 & 0.299 & 0.109 \\
NV~\cite{lombardi2019neural} & 23.93 & 0.784 & 0.276 & 0.080 \\
LLFF~\cite{sitzmann2020implicit} & 23.22 & 0.823 & 0.218 & 0.076 \\ \midrule
NeRF~\cite{mildenhall2020nerf} & 28.65 & 0.856 & 0.206 & 0.047 \\
MipNeRF~\cite{barron2021mip} & \textbf{29.30} & 0.864 & 0.190 & 0.044 \\ \midrule
Aug-NeRF (Ours) & 29.18 & \textbf{0.879} & \textbf{0.173} & \textbf{0.042} \\
\bottomrule
\end{tabular}}
\resizebox{0.9\linewidth}{!}{
\begin{tabular}{l|ccc|c}
\toprule
\multirow{1}{*}{Scene ``\textit{chair}"} & \multirow{1}{*}{PSNR $\uparrow$} & \multirow{1}{*}{SSIM $\uparrow$} & \multirow{1}{*}{LPIPS $\downarrow$} & \multirow{1}{*}{Average $\downarrow$}\\  
\midrule
SRN~\cite{sitzmann2020implicit} & 26.96 & 0.910 & 0.106 & 0.040\\
NV~\cite{lombardi2019neural} & 28.33 & 0.916 & 0.109 & 0.036 \\
LLFF~\cite{sitzmann2020implicit} & 28.72 & 0.948 & 0.064 & 0.027 \\ \midrule
NeRF~\cite{mildenhall2020nerf} & 33.00 & 0.967 & 0.046 & 0.016 \\
MipNeRF~\cite{barron2021mip} & 33.82 & 0.972 & 0.042 & 0.014 \\ \midrule
Aug-NeRF (Ours) & \textbf{33.87} & \textbf{0.972} & \textbf{0.040} & \textbf{0.014} \\
\bottomrule
\end{tabular}}
\vspace{-2mm}
\end{table}

\begin{table}[t]
\caption{Quantitative comparison of Aug-NeRF against NeRF and MipNeRF for novel view synthesis. All models are trained with noisy data. Performance is reported on the \textit{noise-free} test set.}
\label{tab:res_noisy}
\centering
\vspace{-3mm}
\resizebox{0.9\linewidth}{!}{
\begin{tabular}{l|ccc|c}
\toprule
\multirow{1}{*}{``\textit{fern}" + Gaussian Noise} & \multirow{1}{*}{PSNR $\uparrow$} & \multirow{1}{*}{SSIM $\uparrow$} & \multirow{1}{*}{LPIPS $\downarrow$} & \multirow{1}{*}{Average $\downarrow$}\\  
\midrule
NeRF~\cite{mildenhall2020nerf} & 16.95 & 0.451 & 0.535 & 0.200 \\
Aug-NeRF (Ours) & 17.12 & 0.535 & 0.495 & 0.187 \\
\bottomrule
\end{tabular}}
\resizebox{0.9\linewidth}{!}{
\begin{tabular}{l|ccc|c}
\toprule
\multirow{1}{*}{``\textit{fern}" + Shot Noise} & \multirow{1}{*}{PSNR $\uparrow$} & \multirow{1}{*}{SSIM $\uparrow$} & \multirow{1}{*}{LPIPS $\downarrow$} & \multirow{1}{*}{Average $\downarrow$}\\  
\midrule
NeRF~\cite{mildenhall2020nerf} & 15.75 & 0.231 & 0.755 & 0.260 \\
Aug-NeRF (Ours) & 17.00 & 0.495 & 0.485 & 0.190 \\
\bottomrule
\end{tabular}}
\vspace{-3mm}
\end{table}

\begin{figure*}[t]
    \centering
    \vspace{-3mm}
    \includegraphics[width=0.9\linewidth]{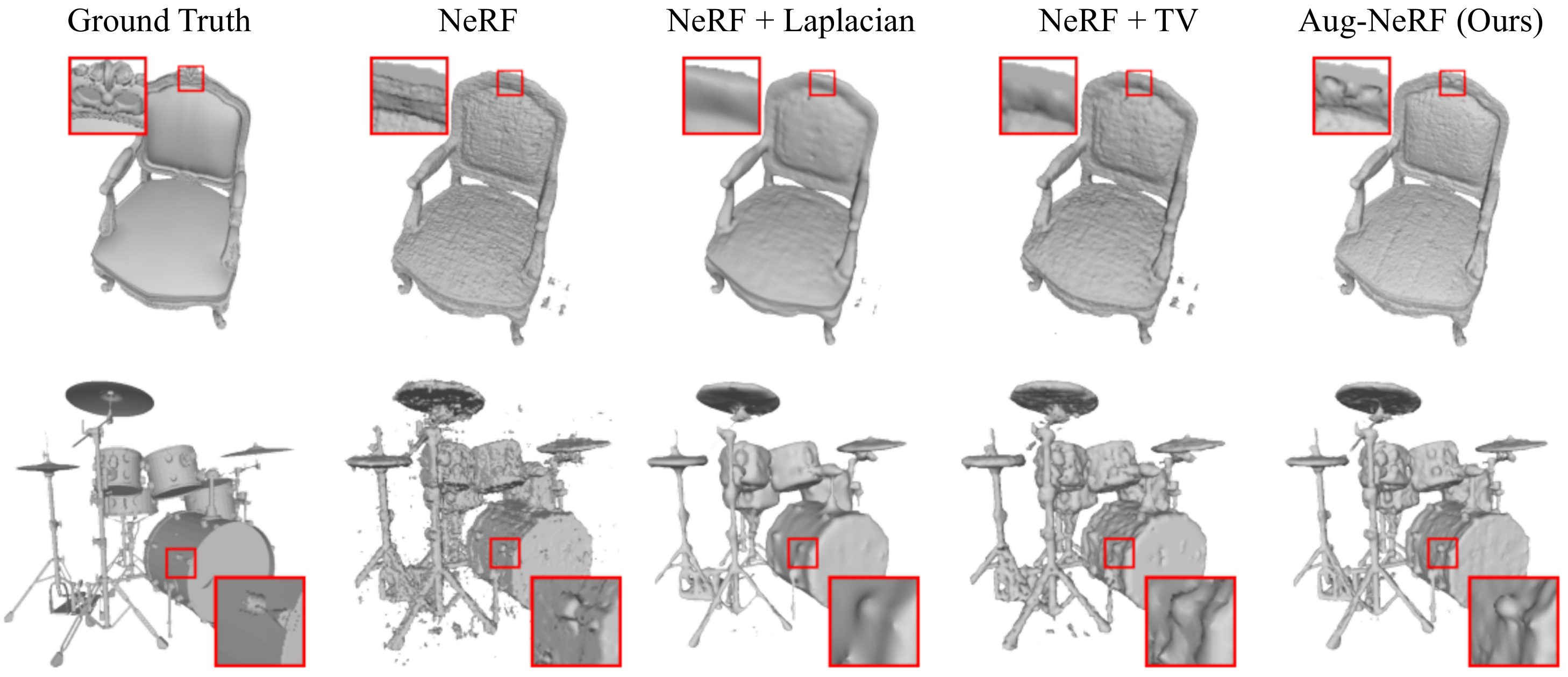}
    \vspace{-3mm}
    \caption{Comparisons of the fitted geometry by NeRF, NeRF with explicit Laplacian and TV regularizers, and our Aug-NeRF. The local zoom-in is placed in the \textcolor{red}{red} box.}
    \vspace{-3mm}
    \label{fig:geometic}
\end{figure*}

\vspace{-1em}
\paragraph{Depth and Geometry visualization.} The learned depth maps and fitted 3D geometries from NeRFs are provided in and~\ref{fig:depth} and Fig.~\ref{fig:geometic}, respectively. The 3D shapes (Fig.~\ref{fig:geometic}) are synthesized by MarchingCube algorithms \cite{lorensen1987marching}. We observe that vanilla NeRF suffers from a serrated surface (which overwhelms the fine details), while traditional TV and Laplacian regularizations tend to excessively smoothen the results. Aug-NeRF reduces noises and improves surface smoothness, in a detail- and geometry-preserving manner. 

\begin{figure}[!ht]
    \centering
    \includegraphics[width=0.99\linewidth]{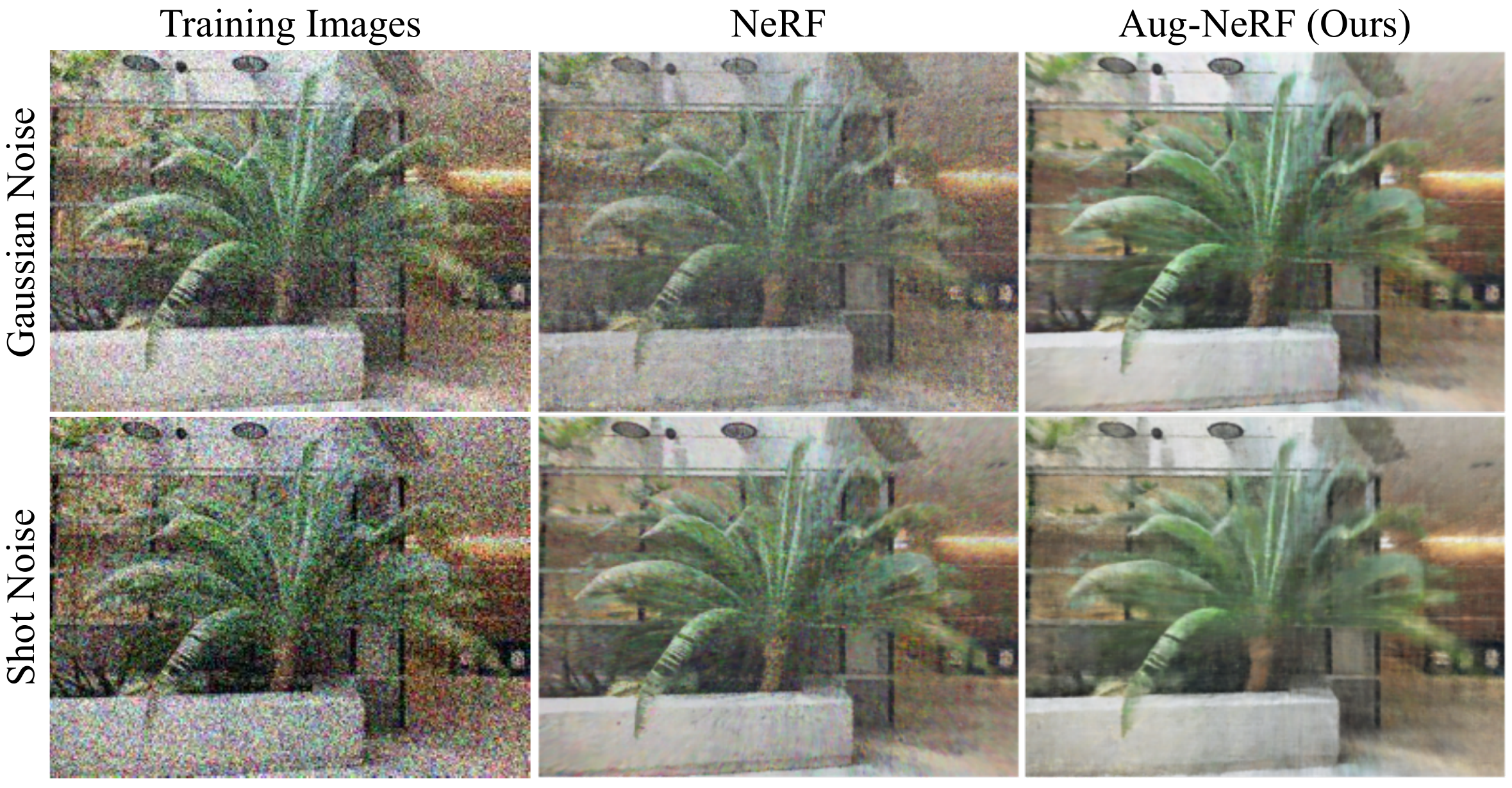}
    \vspace{-3mm}
    \caption{AugNeRF yields superior synthesis results when trained with noisy image supervision. Our investigated image corruptions include Gaussian and shot noises, following the standard in~\cite{hendrycks2019benchmarking}.}
    \label{fig:imgc}
    \vspace{-1em}
\end{figure}

\vspace{-0.7em}
\paragraph{Superior synthesis when trained on noisy data.} As an extra study, we examine Aug-NeRF under supervision images with additive noise corruptions. From Tab.~\ref{tab:res_noisy} and Fig.~\ref{fig:imgc}, compared to the vanilla NeRF, Aug-NeRF shows consistent $6.5\%\sim26.9\%$ average error reductions for both Gaussian and Shot noises, while it substantially improves the visual quality of constructed test views (e.g., much fewer noises in the ``\textit{fern}"). We regard it as an additional bonus from enforcing smooth geometry in NeRF training. 

\subsection{Ablation Study} \label{sec:ablation}

\begin{table}[t]
\caption{Quantitative ablation study of our Aug-NeRF. Input, feature, and output augmentations. denote our proposed coordinate, feature, and pre-rendering output augmentations respectively.}
\label{tab:ablation_all}
\centering
\vspace{-3mm}
\resizebox{0.9\linewidth}{!}{
\begin{tabular}{l|ccc|c}
\toprule
\multirow{1}{*}{Scene ``\textit{fern}"} & \multirow{1}{*}{PSNR $\uparrow$} & \multirow{1}{*}{SSIM $\uparrow$} & \multirow{1}{*}{LPIPS $\downarrow$} & \multirow{1}{*}{Average $\downarrow$}\\  
\midrule
NeRF~\cite{mildenhall2020nerf} & 25.17 & 0.792 & 0.280 & 0.073 \\ \midrule
+ $\ell_1$ Reg. & 25.15 & 0.750 & 0.285 & 0.076 \\ 
+ Lap. Reg. & 24.89  & 0.670 & 0.305 & 0.083\\ 
+ TV Reg. & 26.05 & 0.806 & 0.217 & 0.062\\ 
+ Random Aug. & 25.28 & 0.796 & 0.224 & 0.067 \\
\midrule
+ Input Aug. & 25.30 & 0.797 & 0.251 & 0.069 \\
+ Feature Aug. & 25.39 & 0.787 & 0.243 & 0.069 \\ 
+ Feature \& (Pre-) Output Aug. & 26.32 & 0.810 & 0.199 & 0.059 \\
+ Tri-level Random Noise & 25.36 & 0.802 & 0.205  & 0.064 \\
\midrule
Aug-NeRF (Ours) & 26.51 & 0.830 & 0.168 & 0.054 \\
\bottomrule
\end{tabular}}
\vspace{-5mm}
\end{table}

\vspace{-0.2em}
\paragraph{Multi-level v.s. single-level augmentation.} To compare the effects of robust augmentations at different levels, we conduct step-wise evaluation as: ($i$) NeRF, ($ii$) NeRF + Feature Aug., ($iii$) NeRF + Feature \& Output Aug., $iv$) NeRF + Feature \& Output \& Input coordinates Aug., which is our  complete Aug-NeRF. Tab.~\ref{tab:ablation_all} shows that applying robust augmentation to each level brings extra and complementary generalization gains, among which augmenting the pre-rendering output level makes the biggest difference. 

\vspace{-0.7em}
\paragraph{Worst-case v.s. random perturbations.} One straightforward baseline for Aug-NeRF is to just use random data augmentation. Particularly, we employ random Gaussian noises to both intermediate features and pre-rendering outputs of NeRF\footnote{The vanilla NeRF has already included random noise in coordinates.}. As in Tab.~\ref{tab:ablation_all}, Random Aug. obtains moderate performance boosts for all metrics, but are clearly less obvious than our worst-case perturbations.

\begin{figure}[!ht]
    \centering
    \vspace{-0.5em}
    \includegraphics[width=1\linewidth]{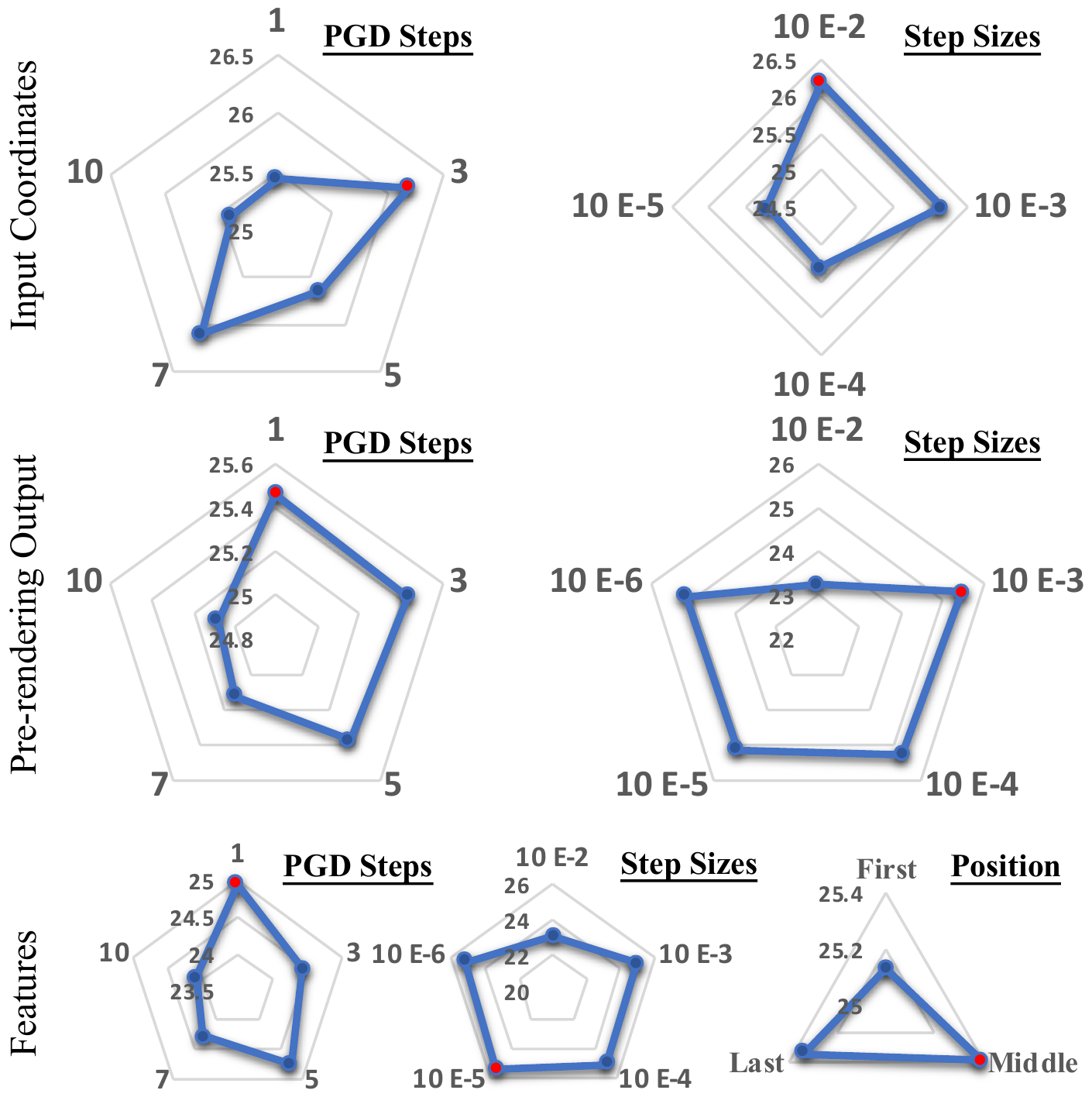}
    \vspace{-4mm}
    \caption{Ablations on the strength and location of all three augmentations. Results are on the test set of LLFF. The PGD step is the number of iterations for generating worse-case perturbations. Step size $\alpha$ controls the strength of crafted perturbations. Position means which layer the intermediate feature augmentation is injected into NeRF's MLP. \textcolor{red}{Red} indicates the top performance.}
    \vspace{-5mm}
    \label{fig:ablation}
\end{figure}

\vspace{-0.7em}
\paragraph{Effects of augmentation strength and location.} The accuracy gains from Aug-NeRF are largely determined by the strength and location of crafted worst-case perturbations. A comprehensive investigation on three levels of augmentations, i.e., input coordinate, intermediates features, and pre-rendering output, are presented in Fig.~\ref{fig:ablation}. When studying one of the factors, we stick to the best configuration for the rest factors. Fig.~\ref{fig:ablation} reveals that: \underline{First}, NeRF gains the most from \{coordinate, features, pre-rendering output\} augmentations with \{PGD-3, PGD-1, PGD-1\} and step size \{$10^{-2}$, $10^{-3}$, $10^{-5}$\}; \underline{Second}, applying generated perturbations to the middle layer of NeRF's MLP contributes the most significantly; \underline{Third}, too strong (e.g., PGD-10) worst-case perturbations may still deteriorate performance. 

\vspace{-0.7em}
\paragraph{Comparison with explicit smooth regularizations.} In contrast to our implicit smooth prior, there exists several explicit smooth regularizations which can be directly plugged into the NeRF pipeline, like \ding{172} $\ell_1$ sparsity Reg. $R_{\ell_1}(\Theta) = \int\nolimits \lvert \sigma_{\Theta}(\Mat{u}) \rvert  d\Mat{u}$; \ding{173} Laplacian Reg. $R_{\mathrm{Lap}}(\Theta) = \int \lvert \Delta \sigma_{\Theta}(\Mat{u}) \rvert d\Mat{u}$; \ding{174} Total Variation (TV) Reg. $R_{\mathrm{TV}}(\Theta) = \int\nolimits \lVert \nabla \sigma_{\Theta}(\Mat{u}) \rVert_2  d\Mat{u}$. As demonstrated in Tab.~\ref{tab:ablation_all}, although hyperparameters are carefully tuned by a grid search, both $\ell_1$ and Laplacian regularizers degrade the performance, as such explicit constraints are often too aggressive and limit the representation flexibility of NeRF. The TV Reg. can lead to positive gains but still largely lags behind our proposals.

\section{Conclusion and Broad Impact}
\vspace{-1mm}
In this paper, we have presented Aug-NeRF that addresses the inherent non-smooth geometries of NeRF. Specifically, based on solid physical grounds, Aug-NeRF seamlessly injects worst-case perturbations into three levels of the NeRF pipeline, leading to substantially improved geometry continuity and generalization ability. Extensive quantitative and qualitative results across diverse scenes validate the effectiveness of our proposals. Moreover, the implicit smooth prior induced by triple-level augmentation enables NeRF to recover scenes from noisy supervision images. One limitation is that we only study additive noises (e.g., Gaussian) for corrupted images. We will extend the investigation to other complicated corruptions.


\clearpage

{\small
\bibliographystyle{ieee_fullname}
\bibliography{Aug-NeRF}
}

\clearpage

\appendix

\renewcommand{\thepage}{A\arabic{page}}  
\renewcommand{\thesection}{A\arabic{section}}   
\renewcommand{\thetable}{A\arabic{table}}   
\renewcommand{\thefigure}{A\arabic{figure}}

\section{More Technical Details} \label{sec:more_technical}

We summarize the detailed procedures of Aug-NeRF in the Algorithm~\ref{alg:pipeline}.

\begin{algorithm}[H]
\caption{The training pipeline of Aug-NeRF. For simplicity, we assume batch size is 1.}
\label{alg:pipeline}
\renewcommand{\algorithmicensure}{\textbf{Initialize:}}
\begin{algorithmic}[1]
\Ensure{Training view images $\Set{I} = \{\Mat{I}_i \in \real^M \}_{i=1}^{N}$ and their associated camera poses $\Set{P} = \{\Mat{\phi}_i \in \real^{3 \times 4}\}_{i=1}^{N}$. Define neural radiance field $F_\Theta(\Mat{p}, \Mat{\theta}) = (g \circ f(\Mat{p}, \Mat{\theta}), h \circ f(\Mat{p})): (\Mat{p},\Mat{\theta}) \mapsto (\Mat{c},\sigma)$ as in Sec. \ref{sec:advs}.}
\State Cast rays for each pixel in each $\Mat{I}_i$ via inverse projection with respect to $\Mat{\phi}_i$, and obtain a set of rays $\Set{R} = \{ (\Mat{o}_i, \Mat{d}_i, \Mat{\theta}_i, \widehat{\Mat{C}}_i) \}_{i=1}^{NM}$.
\While{until convergence}
\State Randomly pick a ray $(\Mat{o}_i, \Mat{d}_i, \Mat{\theta}_i, \widehat{\Mat{C}}_i) \in \Set{R}$
\State Generate adversarial perturbations $\delta_t$, $\Mat{\delta}_{xyz}$, $\Mat{\delta}_\theta$, $\Mat{\delta}_f$, $\Mat{\delta}_c$, $\delta_\sigma$ by solving Eqn. \ref{eqn:min_max} using PGD (Eqn. \ref{eqn:pgd}) within the corresponding search space in Sec. \ref{sec:advs}.
\State \textcolor{gray}{\# Sample points along rays.}
\State (Coarse) Sample $K/2$ depth intervals $t_k$ along the rays uniformly
\State (Fine) Sample $K/2$ depth intervals $t_k$ via proportional to coarse sampled densities \cite{mildenhall2020nerf}.
\For{$k \in \{1, \cdots, K\}$}
\State $t^\dag_k = t_k + \delta_{t,k}, \Mat{\theta^\dag}_{i} = \Mat{\theta}_{i} + \Mat{\delta}_{\theta}$.
\State $\Mat{p}_k = \Mat{o}_i + t_k \Mat{d}_i, \Mat{p^\dag}_k = \Mat{o}_i + t^\dag_k \Mat{d}_i + \Mat{\delta}_{xyz}$
\State $(\Mat{c}_k, \sigma_k) = g \circ f(\Mat{p}_k, \Mat{\theta}_i), h \circ f(\Mat{p}_k)$
\State $\Mat{c^\dag}_k = g(f(\Mat{p^\dag}_k)+\Mat{\delta}_f, \Mat{\theta^\dag}_i) + \Mat{\delta}_c$
\State $\sigma^\dag_k = h(f(\Mat{p^\dag}_k+\Mat{\delta}_f))  + \Mat{\delta}_\sigma$
\EndFor
\State \textcolor{gray}{\# Volumetric rendering.}
\State $(\Delta t_k, \Delta t^\dag_k) = t_k-t_{k-1}, t^\dag_k - t^\dag_{k-1}$
\State $T(k) = \exp\left( -\sum_{l=1}^{k-1} \sigma_{l} \Delta t_l \right)$
\State $T^\dag(k) = \exp\left( -\sum_{l=1}^{k-1} \sigma^\dag_{l} \Delta t^\dag_l \right)$
\State $\Mat{C}_i = \sum_{k=1}^{K} T(k) (1 - \exp(-\sigma_{k} \Delta t_k)) \Mat{c}_{k}$
\State $\Mat{C^\dag}_i = \sum_{k=1}^{K} T^\dag(k) (1 - \exp(-\sigma^\dag_{k} \Delta t^\dag_k)) \Mat{c^\dag}_{k}$
\State \textcolor{gray}{\# Train network.}
\State $\mathcal{L} = \lVert \Mat{C}_i - \widehat{\Mat{C}}_i \rVert_2^2 + \lambda \lVert \Mat{C^\dag}_i - \widehat{\Mat{C}}_i \rVert_2^2$
\State Update network parameter $\Theta$ via $\nabla_{\Theta} \mathcal{L}$.
\EndWhile
\end{algorithmic}
\end{algorithm}

\section{More Experiment Results} \label{sec:more_results}

\begin{figure}[t]
    \centering
    \vspace{-2mm}
    \includegraphics[width=\linewidth]{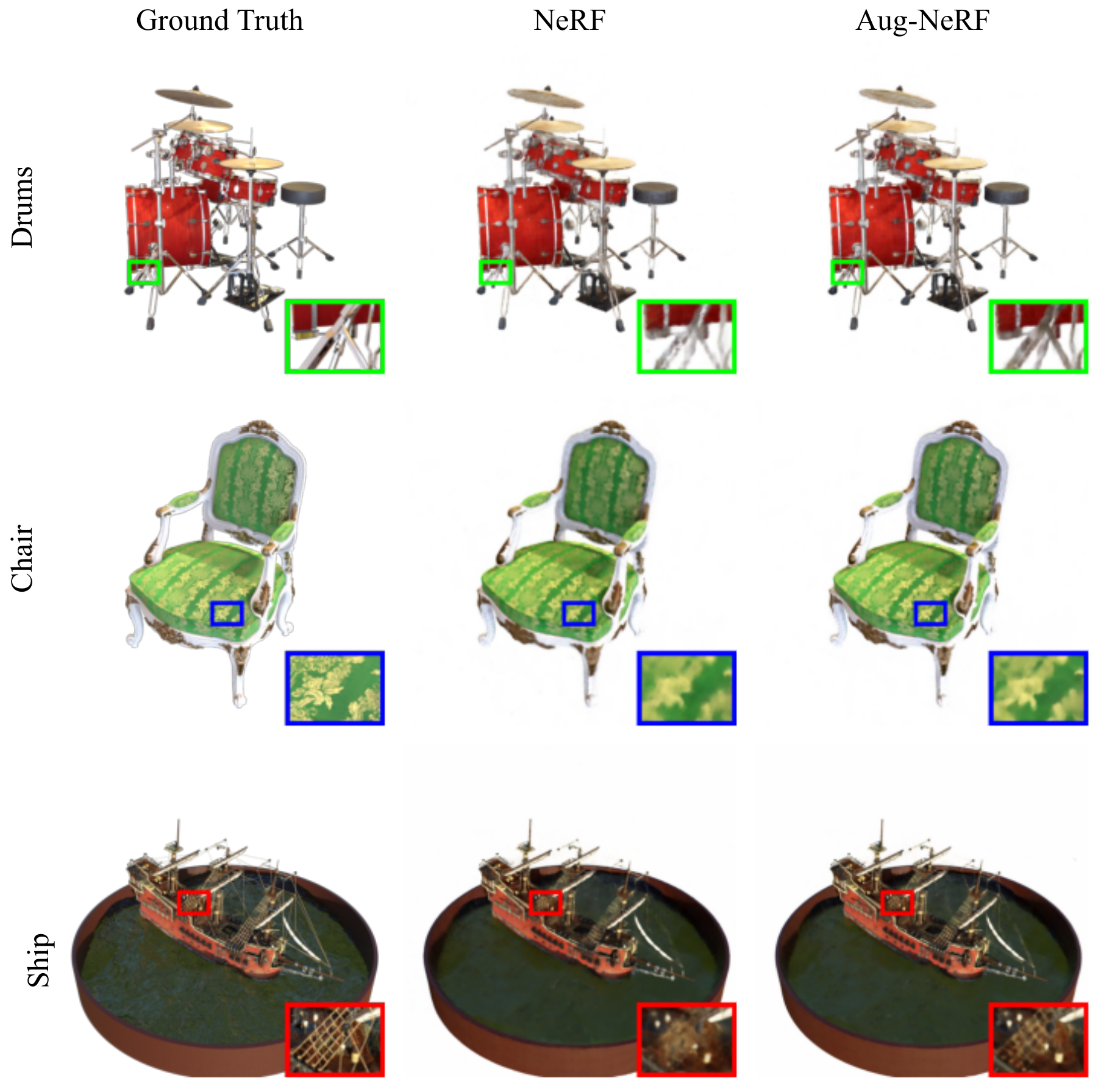}
    \vspace{-4mm}
    \caption{Comparisons on test-set views for scenes from the NeRF-Synthetic 360\textdegree~dataset generated with a physics-based renderer~\cite{mildenhall2020nerf}.}
    \vspace{-5mm}
    \label{fig:res_syn360}
\end{figure}

\begin{figure}[!ht]
    \centering
    \vspace{-2mm}
    \includegraphics[width=\linewidth]{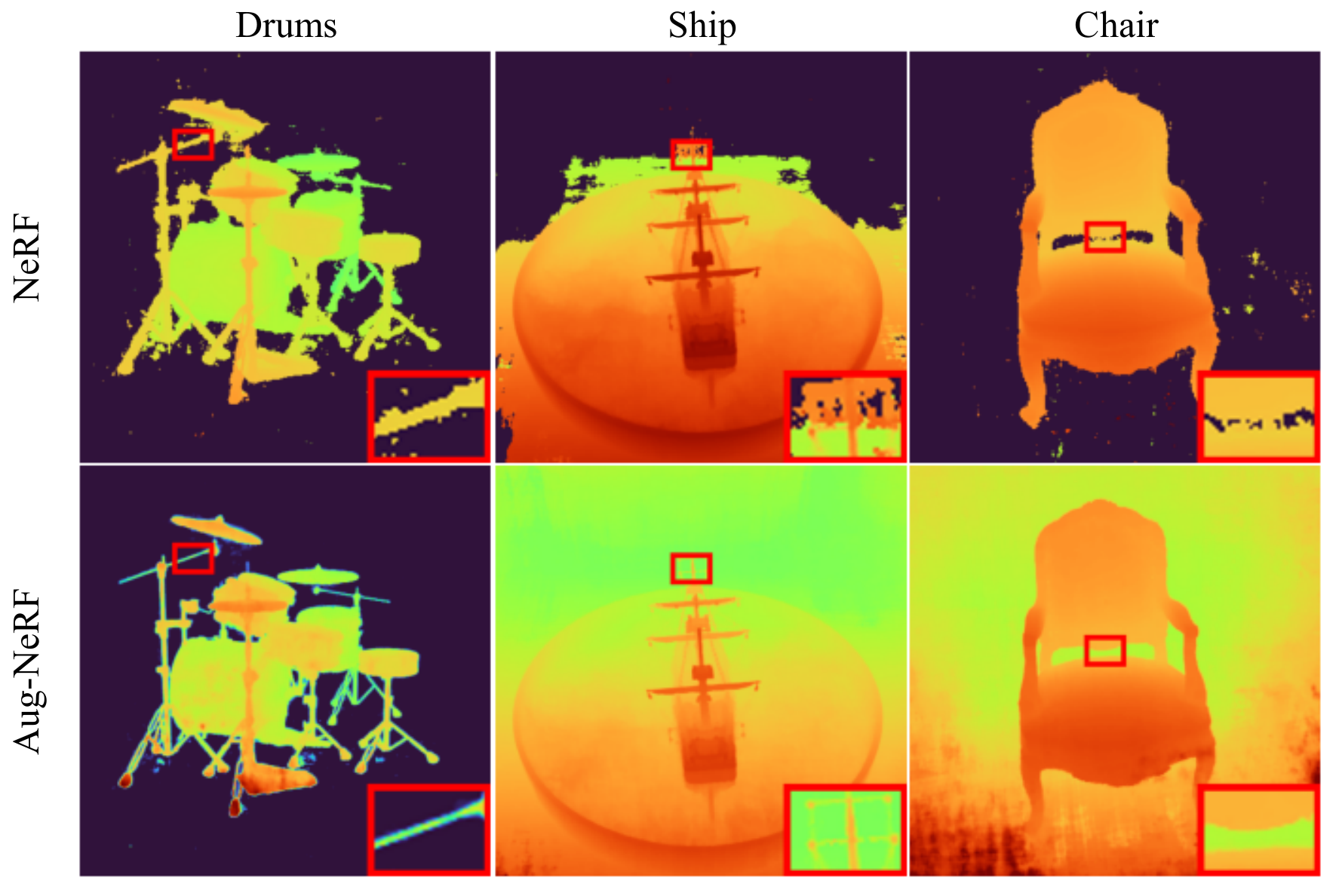}
    \vspace{-3mm}
    \caption{Comparisons of learned depth maps from NeRF and our Aug-NeRF on scenes from NeRF-Synthetic 360\textdegree~dataset.}
    \vspace{-4mm}
    \label{fig:res_syn360_depth}
\end{figure}

\paragraph{Qualitative results on NeRF-Synthetic 360\textdegree~dataset.} We present the constructed test views in Fig.~\ref{fig:res_syn360} and the learned depth maps in Fig.~\ref{fig:res_syn360_depth}. As shown in Fig.~\ref{fig:res_syn360}, we find that the vanilla NeRF fails to capture the fine-grained details of objects, such as the ``\textit{ship net}", while Aug-NeRF demonstrates substantially improved visual qualities.  

In the meantime, from the depth maps in Fig.~\ref{fig:res_syn360_depth}, NeRF baseline suffers from severe noises. On the contrary, Aug-NeRF enjoys much more smooth depth maps, which suggests that our proposed triple-level robust augmentations indeed enhance the NeRF's continuity and generate smooth geometry representations.

\paragraph{Benefits in overfitting vs underfitting cases.} We take the scene ``chair" as an example, and investigate three combinations of different model sizes and data scales as below Tab.~\ref{tab:overfitting}: ($i$) big NeRF (512) on small images ($\frac{1}{2}$ Res.); ($ii$) normal NeRF (256) on small images ($\frac{1}{2}$ Res.); ($iii$) small NeRF (128) on large images (Full Res.). We show that in all settings of overfitting / normal case / underfitting, our proposed augmentations are consistently beneficial.

\begin{table}[!ht]
\centering
\caption{Performance of Aug-NeRF on different backbone and data size combinations.}
\label{tab:overfitting}
\vspace{-3mm}
\resizebox{\linewidth}{!}{
\begin{tabular}{@{}l|l|ccc|c}
\toprule
Setting & Model & \multirow{1}{*}{PSNR $\uparrow$} & \multirow{1}{*}{SSIM $\uparrow$} & \multirow{1}{*}{LPIPS $\downarrow$} & \multirow{1}{*}{Average $\downarrow$}\\
\midrule
\multirow{2}{*}{\begin{tabular}[c]{@{}l@{}} Big NeRF (512) \\ + $\frac{1}{2}$ Res. \end{tabular}} & NeRF & 33.56 & 0.968 & 0.043 & 0.015\\
& Aug-NeRF (Ours) & \textbf{34.26} & \textbf{0.973} & \textbf{0.038} & \textbf{0.013}\\
\midrule
\multirow{2}{*}{\begin{tabular}[c]{@{}l@{}} Big NeRF (256) \\ + $\frac{1}{2}$ Res. \end{tabular}} & NeRF & 33.27 & 0.968 & 0.045 & 0.016 \\
& Aug-NeRF (Ours) & \textbf{33.92} & \textbf{0.971} & \textbf{0.038} & \textbf{0.013} \\
\midrule
\multirow{2}{*}{\begin{tabular}[c]{@{}l@{}} Small NeRF (128) \\ + Full Res. \end{tabular}} & NeRF & 33.00 & 0.967 & 0.046 & 0.016 \\
& Aug-NeRF (Ours) & \textbf{33.86} & \textbf{0.970} & \textbf{0.041} & \textbf{0.014} \\
\bottomrule
\end{tabular}}
\vspace{-2em}
\end{table}%

\paragraph{Geometry extraction.}
To obtain geometric visualization in Fig.~\ref{fig:geometic}, we first query the network $F_\Theta$ with a regular lattice defined over $[-1,1]^3$, and export a discretized density field volume. The absolute voxel size is 2/512. Then we employ marching cube algorithm \cite{lorensen1987marching} provided in UCSF Chimera\footnote{\url{https://www.cgl.ucsf.edu/chimera/}} to  extract the surface. We set the threshold to 25 and 1 for chairs and drums, respectively. The step size is chosen as 1.
In order to numerically assess the quality of reconstructed geometries, we introduce Chamfer Distance (CD) to measure the difference between reconstructed geometries and ground-truth models:
\begin{align*}
d_{CD} = \frac{1}{\lvert \Set{S}_1 \rvert} \sum_{\Mat{x} \in \Set{S}_1} \min_{\Mat{y} \in \Set{S}_2} \lVert \Mat{x} - \Mat{y} \rVert_2 + \frac{1}{\lvert \Set{S}_2 \rvert} \sum_{\Mat{x} \in \Set{S}_2} \min_{\Mat{y} \in \Set{S}_1} \lVert \Mat{x} - \Mat{y} \rVert_2,
\end{align*}
where $\Set{S}_1$ and $\Set{S}_2$ are point sets sampled from the extracted surfaces and ground-truth models, respectively.
On scene chair, our AugNeRF achieves $1.04 \times 10^{-2}$ CD which is 29.25\% lower than vanilla NeRF ($1.47 \times 10^{-2}$).

\paragraph{Different types of noise and inaccurate camera poses.} As shown in Tab.~\ref{tab:res_noisy} and Fig.~\ref{fig:imgc}, we experiment on two kinds of corruptions, i.e., Gaussian and Shot noises. In this paragraph, we add extra results of training with Pepper noise and inaccurate camera poses are collected in below Tab.~\ref{tab:extra_noise}. The results consistently demonstrate the superiority of Aug-NeRF.
We note that our main goal is to endow NeRF with smoothness-aware geometry reconstruction, enhanced generalization to synthesizing unseen views, while the improved tolerance of noisy supervisions is a by-product bonus.

\begin{table}[!ht]
\centering
\caption{Additional results of Aug-NeRF trained on images corrupted by pepper noise and inaccurate camera poses.}
\label{tab:extra_noise}
\vspace{-3mm}
\resizebox{\linewidth}{!}{
\begin{tabular}{@{}l|l|ccc|c}
\toprule
Noise Type & \multirow{1}{*}{``\textit{fern}"} & \multirow{1}{*}{PSNR $\uparrow$} & \multirow{1}{*}{SSIM $\uparrow$} & \multirow{1}{*}{LPIPS $\downarrow$} & \multirow{1}{*}{Average $\downarrow$}\\  
\midrule
\multirow{2}{*}{Pepper Noise} & NeRF & 19.01 & 0.401 & 0.546 & 0.174 \\
& Aug-NeRF (Ours) & \textbf{19.96} & \textbf{0.568} & \textbf{0.403} & \textbf{0.138} \\
\midrule
\multirow{2}{*}{Inaccurate Pose} & NeRF & 12.31 & 0.253 & 0.725 & 0.333 \\
& Aug-NeRF (Ours) & \textbf{13.54} & \textbf{0.365} & \textbf{0.811} & \textbf{0.306} \\
\bottomrule
\end{tabular}}
\vspace{-2em}
\end{table}

\paragraph{Implementation of explicit regularization.}
We investigate three types of explicit regularizations: $\ell_1$ sparsity, total variation (TV), and Laplacian.
The TV regularization is defined as:
\begin{align*}
    R_{\text{TV}}(\Theta) = \int_{[-1,1]^3} \lvert \nabla_{\Mat{x}} \sigma_{\Theta}(\Mat{x})  \rvert d\Mat{x},
\end{align*}
where $\sigma_\Theta$ denotes the density branch of the function $F_\Theta$. However, evaluating this integral is implausible. Instead, we discretize the integral interval into regular grids and conducting quadrature rule for estimating TV regularization: 
\begin{align*}
    R_{\text{TV}}(\Theta) = \sum_{i=1}^{N} \sum_{j=1}^{N} \sum_{k=1}^{N} \lvert \nabla_{\Mat{x}} \sigma_{\Theta}(\delta i, \delta j, \delta k) \rvert \delta^3,
\end{align*}
where we utilize auto-differentiation provided in PyTorch Library \cite{paszke2017automatic} to calculate $\nabla_{\Mat{x}} \sigma_{\Theta}$.
Similarly, we can approximate $\ell_1$ sparsity and Laplacian regularization by:
\begin{align*}
    R_{\ell_1}(\Theta) &= \int_{[-1,1]^3} \lvert \sigma_{\Theta}(\Mat{x})  \rvert d\Mat{x} \\
    &\approx \sum_{i=1}^{N} \sum_{j=1}^{N} \sum_{k=1}^{N} \lvert \sigma_{\Theta}(\delta i, \delta j, \delta k) \rvert \delta^3  \\
    R_{\text{Lap}}(\Theta) &= \int_{[-1,1]^3} \lvert \Delta_{\Mat{x}} \sigma_{\Theta}(\Mat{x})  \rvert d\Mat{x} \\
    &\approx \sum_{i=1}^{N} \sum_{j=1}^{N} \sum_{k=1}^{N} \lvert \Delta_{\Mat{x}} \sigma_{\Theta}(\delta i, \delta j, \delta k) \rvert \delta^3  \\
\end{align*}
where $\Delta = \operatorname{div} \cdot \nabla$ denotes the Laplacian operator.

\end{document}